\documentclass{article}

\usepackage[english]{babel}
\usepackage{orcidlink}

\usepackage[letterpaper,top=2cm,bottom=2cm,left=3cm,right=3cm,marginparwidth=1.75cm]{geometry}

\usepackage{listings}
\usepackage[english]{babel}
\usepackage{booktabs}
\usepackage{siunitx}
\usepackage{graphicx}
\usepackage{multirow}
\usepackage{subcaption}
\usepackage{float}
\usepackage{pgfplots}
\usepgfplotslibrary{statistics,colorbrewer}

\usepackage[most]{tcolorbox}
\usepackage{xcolor}

\newtcolorbox{commentbox}[1][]{ colback=red!5!white, 
colframe=red!75!black, 
title=@VP, fonttitle=\bfseries, #1 }

\usepackage{pgfplots}
\pgfplotsset{compat=1.18} 

\pgfplotsset{
/pgfplots/matplotlib style/.style={ axis background/.style={fill=white}, axis line style={black}, 
tick style={black}, 
grid=both, 
grid style={gray!30, dashed}, 
xlabel style={font=\small\sffamily}, ylabel style={font=\small\sffamily}, legend style={ draw=none, 
fill=white, 
font=\small\sffamily, cells={anchor=west}, }, every axis plot/.append style={line width=1pt}, tick label style={font=\small\sffamily}, }
}

\usepackage[ruled, linesnumbered]{algorithm2e}
\SetKwProg{Fn}{function}{ is}{end}
\SetKwComment{Comment}{/* }{ */}

\usepackage{makecell}

\usepackage{amsmath}

\usepackage{graphicx}

\usepackage{pgfkeys}
\pgfkeys{
/fitnessargs/.cd,
xlabel/.initial=Generation,
ylabel/.initial=Fitness Value,
legend/.initial=Best fitness,
color/.initial={rgb,255:red,31; green,119; blue,180},
}

\NewDocumentCommand{\fitnessplot}{ O{} m }{
\pgfkeys{/fitnessargs/.cd, #1}

\begin{tikzpicture}
\begin{axis}[
    matplotlib style, 
    xmin=0,
    yticklabel style={/pgf/number format/.cd, fixed, precision=3 },
    width=0.9\linewidth,
    height=6cm,
    xlabel=\pgfkeysvalueof{/fitnessargs/xlabel},
    ylabel=\pgfkeysvalueof{/fitnessargs/ylabel},
    legend pos=south east,
]
    \addplot [ color={\pgfkeysvalueof{/fitnessargs/color}}, 
    mark=none, ] table [x=step, y=value, col sep=comma] {#2};

    \addlegendentry{\pgfkeysvalueof{/fitnessargs/legend}}
\end{axis}
\end{tikzpicture}%
}

\newcommand{\quotemarks}[1]{``#1''}

\usepackage{amsmath, amssymb, amsfonts}
\usepackage{algorithmic}
\usepackage{graphicx}
\usepackage{textcomp}
\usepackage{xcolor}

\begin{document}

\title{\Huge \bfseries Freeze and Conquer:\\
Reusable Ansatz for Solving the Traveling Salesman Problem}

\author{
  Fabrizio Fagiolo\textsuperscript{1,*}\,\orcidlink{0009-0003-0390-7855}\thanks{These authors contributed equally.}
  \and
  Nicolò Vescera\textsuperscript{2,*}\,\orcidlink{0009-0009-0655-2168}
}

\date{
  \textsuperscript{1}\,National Research Council (CNR), Rome, Italy \\
  \textsuperscript{2}\,Università degli Studi di Perugia, Perugia, Italy \\
  \vspace{0.5em}
  \texttt{fabrizio.fagiolo@istc.cnr.it} \quad
  \texttt{nicolo.vescera@collaboratori.unipg.it}
}

\maketitle

\begin{abstract}
    In this paper we present a variational algorithm for the Traveling Salesman
    Problem (TSP) that combines (i) a compact encoding of permutations,
    which reduces the qubit requirement too, (ii) an optimize-freeze-reuse strategy:
    where the circuit topology (\quotemarks{Ansatz}) is first optimized on a
    training instance by Simulated Annealing (SA), then \quotemarks{frozen} and
    re-used on novel instances, limited to a rapid re-optimization of only
    the circuit parameters. This pipeline eliminates costly structural research
    in testing, making the procedure immediately implementable on NISQ
    hardware.


    On a set of $40$ randomly generated symmetric instances that span $4 - 7$
    cities, the resulting Ansatz achieves an average optimal trip sampling
    probability of $100\%$ for 4 city cases, $90\%$ for 5 city cases and $80\%$
    for 6 city cases. With 7 cities the success rate drops markedly to an
    average of $\sim 20\%$, revealing the onset of scalability limitations of
    the proposed method.

    The results show robust generalization ability for moderate problem sizes
    and indicate how freezing the Ansatz can dramatically reduce time-to-solution
    without degrading solution quality. The paper also discusses scalability
    limitations, the impact of \quotemarks{warm-start} initialization of
    parameters, and prospects for extension to more complex problems, such
    as Vehicle Routing and Job-Shop Scheduling.
\end{abstract}

\maketitle

\section{Introduction}
\label{introduction}

Solving combinatorial optimization problems is among the most prominent applications of quantum computing. 
Two main paradigms have emerged to tackle such problems: quantum annealers, which are specialized hardware devices designed to minimize a cost Hamiltonian, and gate-based quantum processors, which implement Variational Quantum Algorithms (VQAs) such as the Variational Quantum Eigensolver (VQE) and the Quantum Approximate Optimization Algorithm (QAOA).
However, when dealing with NP-hard combinatorial problems such as the Traveling Salesman Problem (TSP), two recurring bottlenecks emerge in quantum technology: (i) the use of $\mathcal{O}(n^2)$ qubits in standard QUBO/HUBO encodings, which require heavy penalty terms to enforce feasibility, and (ii) deep, fixed circuit layouts that are ill-suited to the constraints of NISQ devices.

In this work, we use the Variational Quantum Eigensolver (VQE), where one prepares a parameterized Ansatz $\mathcal{A}$, $|\psi(\boldsymbol{\theta})\rangle$, and adjusts the parameter vector $\boldsymbol{\theta}$ so as to minimize
\[
E(\boldsymbol{\theta}) \;=\; \langle \psi(\boldsymbol{\theta}) \,|\, \hat{H} \,|\, \psi(\boldsymbol{\theta}) \rangle,
\]
thereby obtaining an approximation of the ground–state energy $E_{\min}$ of
$\hat{H}$ \cite{peruzzo2014variational,Kandala_2017}. The shallow depth and flexibility of the circuit make VQE particularly suitable for NISQ hardware
\cite{preskill2018quantum,bharti2022noisy}.

Another approach to solving the combinatorial optimization problem is the Quantum Approximate Optimization Algorithm (QAOA) that alternates
$p$ layers of two unitaries,
\[
\begin{aligned}
U_{C}(\gamma_{k}) & = e^{-i\gamma_k \hat{H}_C}, \\
U_{B}(\beta_{k})  & = e^{-i\beta_k \hat{H}_B},
\end{aligned}
\qquad k = 1,\dots,p,\qquad \hat{H}_{B}= \sum_{j=1}^{n} X_{j},
\]
to construct the variational state.
\[
|\psi_{p}(\boldsymbol{\gamma},\boldsymbol{\beta})\rangle
= \Bigl(\prod_{k=1}^{p} U_{B}(\beta_{k})\,U_{C}(\gamma_{k})\Bigr)\, |+\rangle^{\otimes n},
\]
where $\boldsymbol{\gamma}=(\gamma_{1},\ldots,\gamma_{p})$ and
$\boldsymbol{\beta}=(\beta_{1},\ldots,\beta_{p})$ are optimized to minimize $\langle \hat{H}_C \rangle$ \cite{farhi2014quantum,albash2018adiabatic}.
Because the circuit explicitly mirrors $\hat{H}_C$, the exploration of the Hilbert space is biased toward states that are relevant to the target optimization problem.

The Traveling Salesman Problem serves as our case study in this work.
Let $V=\{v_{1},\dots,v_{n}\}$ be a set of $n$ cities and let $c: V\times V \to \mathbb{R}_{\ge 0}$ be a non‑negative cost function with 
$c_{ij}=c(v_i,v_j)$. A \emph{tour} is a permutation $\pi \in S_n$ of $\{1,\dots,n\}$ that induces the closed walk  $\,v_{\pi(1)} \to v_{\pi(2)} \to \cdots \to v_{\pi(n)} \to v_{\pi(1)}\,$  with total cost
\[
C(\pi) \;=\; \sum_{k=1}^{n} c_{\pi(k),\,\pi(k+1)}, 
\qquad \text{with the convention } \pi(n+1)=\pi(1).
\]
The optimization task is to find $\pi^\ast \in S_n$ by minimizing $C(\pi)$ over all $n!$ permutations. 
Computing $\pi^\ast$ is NP-hard and exact methods quickly become impractical as $n$ grows.

In the VQE/QAOA formulations of the TSP, one of the first fundamental design choices concerns the encoding of permutations in qubits.
Instead of QUBO/HUBO encodings, we adopt compact representations such as Lehmer encoding or permutation encoding \cite{permutation_encoding} that reduce the number of qubits needed to $\mathcal{O}(n \log n)$.
From this encoding, we develop a two-step workflow inspired by Machine Learning (ML) methods.  
During the training phase, a Simulated Annealing (SA) procedure jointly explores and learns the circuit topology and its parameters. 
In the subsequent test phase, the learned topology is frozen and the parameters are only slightly re-optimized, allowing a direct evaluation of the generalization ability.

Rather than deriving a Hamiltonian of the TSP in closed form, we estimate the objective value by sampling: at each measurement, we translate the outcome into a tour, compute its classical cost, and use the average cost as an estimate of the expected energy; 
instead, the solution is the minimum cost tour among those generated by the circuit.
This allows us to perform variational optimization without explicitly writing the Hamiltonian.

The remainder of the paper is organized as follows. 
Section \ref{related_work} describes the related work. 
Section \ref{proposed_solution} introduces the scheme of the proposed approach. 
Section \ref{experiments} describes and comments on the experimental results. 
Section \ref{conclusions_and_future} ends the paper by providing
some concluding remarks as well as possible future research lines.

\section{Related work}
\label{related_work}

In the Variational Quantum Algorithm (VQA), $\mathcal{A}$ defines the set of
states reachable by the circuit, effectively delimiting the space of explorable
solutions. In the chemical-quantum domain, for example, it has already been
shown that the choice of $\mathcal{A}$ from the hardware efficient circuits
of Kandala et al.\cite{Kandala_2017} decisively affects the trade-off
between circuit depth and accuracy of results.

The same is true for combinatorial problems such as TSP. Venkat et al.
\cite{differernt-architectures} show that both the architecture (annealer versus
gate-based) and the shape of $\mathcal{A}$ decisively affect solution
quality and computation time. Murhaf et al.\cite{alwir}, exploiting D-Wave's
Pegasus topology and a QUBO reformulation that eliminates the initial node, achieve
better tours than D-Wave's \quotemarks{standard} TSP solver and arrive at 20
cities with fewer qubits. In contrast, the unoptimized approach of Jain et
al.\cite{jain2021tsp} stops at 8 cities without showing competitive
performance.

Despite the limited number of qubits, gate-based computers can still hold their
own against annealers through targeted encoding and $\mathcal{A}$. Ruan et
al.\cite{Ruan} integrate TSP constraints into the QAOA mixer, halve the qubits
needed and double the probability of finding the optimal tour. Qian et al.\cite{Qian}
test three alternative mixers, improved edge-based coding and layered learning,
achieving the best trade-off between depth and quality even in the presence
of realistic noise.

Finally, Meng Shi et al.\cite{Shi2025} propose APIs that automatically transform
any problem on graphs (including TSP) into circuits solved by VQA, achieving
with not too many qubits solutions equal to or better than those of
specialized algorithms.

The most recent literature points out that, in VQA, it is not only the shape
of $\mathcal{A}$ that determines performance, but also how parameters are initialized
and updated.

Egger et al.\cite{Egger2021warmstartingquantum} introduce the warm-starting quantum
optimization paradigm, showing that the choice of initial parameters
obtained from a classical relaxation accelerates convergence without
compromising quality guarantees. In a complementary study, Montañez-Barrera
et al.\cite{MontaEzBarrera2025} show that parameters optimized on small
instances of QAOA can be successfully transferred to larger instances, and even
to other combinatorial problems, while maintaining high accuracy of the
solutions. Similarly, with Meta-VQE, Cervera-Lierta et al.\cite{CerveraLierta2021}
train a one time set of parameters that quickly adapts to different Hamiltonian
configurations, reducing optimization time by orders of magnitude.

Moving in this direction, we propose a freeze \& reuse strategy for the TSP:
we design a compact $\mathcal{A}$ that is capable of finding good solutions,
freeze its optimal structure, and, for each new instance, optimize only the
variational parameters. This eliminates structural search, cutting
resolution time without loss of quality, and avoids any hardware reconfiguration,
making the procedure immediately implementable on NISQ devices. $\mathcal{A}$
thus becomes a reusable resource that significantly reduces TSP time-to-solution.

\section{Proposed solution}
\label{proposed_solution}

\begin{algorithm}
    [ht]
    \caption{Decoding permutation procedure}
    \label{alg:permutation}

    \Fn{\texttt{PermutationDecoding}($\mathcal{P}$)}{ $R \gets [1,2,\dots,n]$\; \For{$i \gets n$ \textbf{downto} $1$}{ $j \gets 1+(\mathcal{P}\mod i)$\; $\pi(n+1-i) \gets R_{j}$\; remove $R_{j}$ from $R$\; $L \gets \lfloor \mathcal{P}/ i \rfloor$\; } \textbf{return} $\pi$\; }
\end{algorithm}

The proposed method introduces two key innovations: the dynamic optimization
of an $\mathcal{A}$ tailored to a specific instance, and the reuse of the same
optimized $\mathcal{A}$ on previously unseen instances. Unlike most existing
studies that rely on a fixed circuit topology, we co-optimize the $\mathcal{A}$
layout itself to better suit the problem instance. This is accomplished
through SA, which iteratively explores the space represented by rotation layer
and entanglement patterns.

We decides to use an efficient encoding mechanism to represent our solution. 
As shown by \cite{permutation_encoding}, permutation encoding provides a
more compact representation of a tour. Specifically, given a permutation
$\pi$ of $n$ cities, we encode it as a single integer $P_{\pi}$ in the range
$[0,.., n!-1]$. The original permutation can be recovered using the decoding
procedure outlined in Algorithm \ref{alg:permutation}, which requires only $\mathcal{O}
(n)$ operations, making it highly scalable even for large $n$.

The approach we propose is summarized in Algorithm \ref{alg:main}. It consists
of two main components: (i) optimizing $\mathcal{A}$ by calling SA with one
of the TSP instances $t_{0}\in \mathcal{T}$ (ii) \quotemarks{freezing} the best
found solution's structure and testing it on all remaining instances $\{\mathcal{T}
\setminus t_{0}\}$. This procedure is inspired by the concept of \textit{training}
and \textit{testing} phases from Machine Learning (ML). Similarly to ML, we
\textit{train} $\mathcal{A}$ using a training set $\{t_{0}\in \mathcal{T}\}$,
and then \textit{test} it using a testing set $\{\mathcal{T}\setminus t_{0}\}$.

\begin{algorithm}
    [ht]
    \caption{Optimize, Freeze and Reuse}
    \label{alg:main} \KwData{set of TSP instances $\mathcal{T}$} \KwResult{Optimized and Tested $\mathcal{A}$ (with parameters)}

    generate random $\mathcal{A}$\; use SA to optimize $\mathcal{A}$ using $t
    _{0}\in \mathcal{T}$ as benchmark\; freeze $\mathcal{A}$ shape\; use $\mathcal{A}$
    to solve TSP problems in $\{\mathcal{T}\setminus t_{0}\}$
\end{algorithm}

\subsection{Ansatz Optimization}

The first key component of our approach is utilizing the SA optimization algorithm
to find the optimal $\mathcal{A}_{best}$ for the TSP problem. We choose this
method due to its capability of discovering an approximate global optimum in
a large search space, avoiding local optima. Instead of employing a
population-based approach, we opted for a single-individual algorithm as it can
allocate all available resources towards fully exploring and refining one solution,
thereby accelerating convergence and enabling more profound local search. Our
SA implementation is detailed in Algorithm \ref{alg:SA}.

Selecting an appropriate $\mathcal{A}$ size is challenging because excessively
large circuits incur high computational costs, while undersized ones produce
suboptimal results. To tackle this issue, we partition our $\mathcal{A}$
into 5 distinct blocks, each of which can be either a \textit{Rotation block}
or an \textit{Entanglement block}. Rotation blocks consist of rotation gates
($R_{x}$, $R_{y}$, $R_{z}$), whereas Entanglement blocks are constructed using
$CX$ gates (see \cite{qiskit}). The first block of $\mathcal{A}$ is always a
rotation block, followed by an entanglement block, and this alternating
pattern continues until the final block, which is also a rotation block.
Within each rotation block, every gate has an independent parameter $\Theta_{i}
[j]$, where $i$ denotes the block index and $j$ represents the qubit index.

\begin{equation*}
    \begin{split}
        &\mathcal{A}= \ <R_{block}, CX_{block}, R_{block}, CX_{block}, R_{block}
        >, \\
        &R_{block}\in \{R_{x}, R_{y}, R_{z}\}\\
        &CX_{block}\in \{\text{linear, reverse linear, full, circular, SCA}\}
    \end{split}
\end{equation*}
Let us take into account the following example:

\begin{equation}
    \label{eq:ansatzexample}\mathcal{A}= \ <R_{z}, linear, R_{z}, reverse \ l
    inear, R_{y}>
\end{equation}

\begin{figure}[ht]
    \centering
    \includegraphics[width=1.0\linewidth]{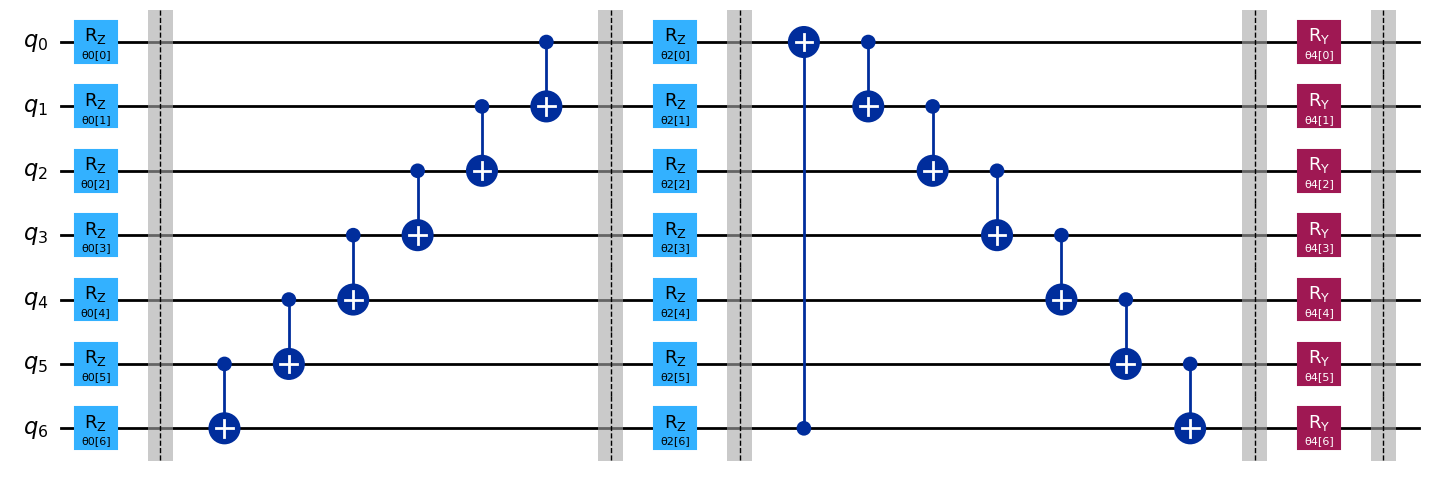}
    \caption{Quantum Circuit associated to example in (\ref{eq:ansatzexample})}
    \label{fig:ansatzexample}
\end{figure}

This $\mathcal{A}$ is represented as a quantum circuit in Figure
\ref{fig:ansatzexample}. The first block consists of $R_{z}$ gates, one for
each qubit, forming a rotation layer. This is followed by an entanglement
block composed of linearly arranged $CX$ gates. The pattern alternates between
rotation and entanglement blocks, ending with a final rotation block of
$R_{y}$ gates.

\begin{algorithm}
    \caption{SA for Ansatz Optimization}
    \label{alg:SA}

    \KwData{TSP instance $t_{0}\in \mathcal{T}$} \KwResult{Optimized $\mathcal{A}_{best}$ (with parameters)}
    $T \gets 1.0$\; $CoolingRate \gets 0.999$\;
    $T_{min}\gets 1 \times 10^{-3}$\; $Iter_{max}\gets 500$\;
    $iteration \gets 0$\;

    $\mathcal{A}_{best}\gets$ generate a random Ansatz\; $\mathcal{A}_{current}
    \gets \mathcal{A}_{best}$\;

    \While{$T > T_{min}$ \textbf{and} $iteration < Iter_{max}$}{ \label{alg:SA:line:termcond} $\mathcal{A}_{neighbor}\gets$ \texttt{Perturbate}($\mathcal{A}_{current}$)\; \label{alg:SA:line:perturbate} $\Delta \text{E}\gets \text{\texttt{Fitness}}(\mathcal{A}_{neighbor}) - \text{\texttt{Fitness}}(\mathcal{A}_{current})$\;



    \If{$\Delta \text{E}> 0$ \textbf{or} $\text{\texttt{RandomNumber}}() < e^{\frac{\Delta \text{E}}{\text{T}}}$} { \label{alg:SA:line:acceptcond} $\mathcal{A}_{current}\gets \mathcal{A}_{neighbor}$\; \If{$\text{\texttt{Fitness}}(\mathcal{A}_{current}) > \text{\texttt{Fitness}}(\mathcal{A}_{best})$}{ $\mathcal{A}_{best}\gets \mathcal{A}_{current}$\; } }

    $T \gets T \times CoolingRate$\; $iteration \gets iteration + 1$\; }
\end{algorithm}

The perturbations of $\mathcal{A}_{current}$, implemented by the \texttt{Perturbate}
function (line \ref{alg:SA:line:perturbate}), involve randomly selecting a block
$b \in \mathcal{A}_{current}$ and replacing it with another distinct, valid
block. This process modifies $\mathcal{A}_{current}$ to generate $\mathcal{A}
_{neighbor}$ while ensuring compliance with the previously defined rules.

The fitness value of a given $\mathcal{A}$ is determined by executing its
quantum circuit. The evaluation proceeds as follows:
\begin{enumerate}
    \item \textbf{Initial Sampling}: We generate $100$ random parameter vectors
        with angles in $[0, 2\pi]$ and compute their corresponding energy values.

    \item \textbf{Selection}: The 10 vectors yielding the lowest energies are
        selected as initial points for optimization.

    \item \textbf{Optimization}: Using the Powell algorithm \cite{powell},
        we refine these parameters to minimize the energy further.
\end{enumerate}

The Powell algorithm was chosen after extensive comparisons with both gradient-based
and gradient-free optimizers, as it demonstrated superior effectiveness and
reliability in our experiments. After optimization, we execute the circuit
$n$ times using the refined parameters and record the outcomes. The fitness value
is then computed as the empirical probability of observing the optimal permutation,
defined as:

\begin{equation}
    Fitness = \frac{\text{Number of optimal permutation observed}}{n}
\end{equation}

This entire evaluation process is encapsulated in the \texttt{Fitness}
procedure (Algorithm \ref{alg:fitenss}). The acceptance condition for replacing
$\mathcal{A}_{current}$ with $\mathcal{A}_{neighbor}$ is implemented at line
\ref{alg:SA:line:acceptcond} of Algorithm \ref{alg:SA} using the Metropolis criterion
\cite{metropolis}. As a fundamental component of SA, this criterion enables an
effective balance between: \textit{exploration} of new solutions and \textit{exploitation}
of current solutions while maintaining the ability to escape local optima
and progressively converge toward global optima in complex search spaces.

\begin{algorithm}
    [ht]

    \caption{Fitness computation}
    \label{alg:fitenss}

    \Fn{\texttt{Fitness}($\mathcal{A}$, $T$: int) $\rightarrow$ float}{ execute \texttt{RunVQE}($\mathcal{A}$) $T$-times and store the results\; $f \gets$ compute the average value over stored results\; \textbf{return} $f$\; }

    \Fn{\texttt{RunVQE}($\mathcal{A}$) $\rightarrow$ list[float]}{ $S \gets$ generate 100 random vector in range $[0, 2\pi]$\; \For{$s \in S$}{ compute $s$ energy\; } $S_{start}\gets$ find first 10 vectors with lowest energy\; $results \gets$ empty list\; \For{$n$ times}{
    $params \gets$ \texttt{Powell}($\mathcal{A}, S_{start}$)\; apply $params$ to $\mathcal{A}$\; $res \gets$ execute VQE using $\mathcal{A}$ ($N=1024$)\; 
    $P_{opt}\gets \frac{\text{number of correct solution found}}{N}$\; add $P_{opt}$ to $results$\; }

    \textbf{return} $results$\; }
\end{algorithm}

\subsection{Reusing the Optimized Ansatz}

%

Upon completion of the SA algorithm, we obtain the optimal
$\mathcal{A}_{best}$ and its associated parameters for the instance
$t_{0}\in \mathcal{T}$. The next step is to apply the freezing procedure:
$\mathcal{A}$ topology is fixed (no further structural changes are allowed)
and it is used to solve all other instances in
$\{\mathcal{T}\setminus t_{0}\}$. Our \quotemarks{freezing} procedure is
listed in Algorithm \ref{alg:freeze}.

The core advantage of this approach lies in its ability to reuse a pre-optimized
$\mathcal{A}$, both in structure and parameters, to solve new problem
instances without restarting the full optimization process. The method simply
requires local re-optimization via Powell's algorithm (line
\ref{alg:freeze:line:powell}), using the previously identified optimal
parameters from SA as starting points, to efficiently derive a new optimized
parameter set specifically tailored to each target instance.

\begin{algorithm}
    [ht]
    \caption{Freezing procedure}
    \label{alg:freeze}

    \KwData{$\mathcal{A}_{best}$, $OptParams_{\mathcal{A}_{best}}$, $\{\mathcal{T}\setminus t_{0}\}$}
    \KwResult{$P_{opt}$ foreach $t_{i}\in \{\mathcal{T}\setminus t_{0}\}$}

    $results \gets$ empty list\; \For{$t_{i}\in \{\mathcal{T}\setminus t_{0}\}$}{ $params \gets$ \texttt{Powell}($\mathcal{A}_{best}, OptParams_{\mathcal{A}_{best}}$)\; \label{alg:freeze:line:powell} apply $params$ to $\mathcal{A}_{best}$\; $res \gets$ execute VQE using $\mathcal{A}_{best}$ ($N = 1024$)\; $P_{opt}\gets \frac{\text{number of correct solution found}}{N}$\; add $P_{opt}$ to $results$\; }
\end{algorithm}

\section{Experiments}
\label{experiments}

This section presents the experimental results of applying our VQA approach to
the TSP.

\subsection{Problem setup}
\label{problem-setup}

We generate TSP instances using a controlled procedure, where each instance is
represented by a symmetric $n \times n$ distance matrix, with $n$ denoting
the number of cities. Distances between cities are randomly sampled as integers
within a predefined range (e.g., $1$ to $100$), while self-distances are set
to 0 to ensure validity. This generation method enables the efficient construction
of instances for testing and comparing algorithms across varying levels of complexity
and problem size.

We generated $44$ TSP instances in total, 11 for each city dimension ($4$,$5$,$6$,$7$).
For each dimension, instance $0$ is reserved for the training of
$\mathcal{A}$, while instances $1-10$ form the test set used to evaluate the
generalization of the algorithm.

This configuration also allows the optimal tour cost to be calculated through
an exhaustive enumeration of all city permutations by a classical method,
providing an exact reference solution. This exact value serves as a benchmark
to assess the performance of our VQA approach, by comparing the cost of the
generated tours against the known optimum. The number of qubits and $\mathcal{A}$
parameters required for each instance size is reported in Table \ref{tab:qubits_parameters}.

\begin{table}[ht]
    \centering
    \caption{Qubit and Ansatz parameter count required for different city
    sizes.}
    \label{tab:qubits_parameters}
    \begin{tabular}{ccc}
        \toprule \textbf{City nodes} & \textbf{Qubits used} & \textbf{Ansatz parameters} \\
        \midrule $4$                 & 5                    & 15                         \\
        $5$                          & 7                    & 21                         \\
        $6$                          & 10                   & 30                         \\
        $7$                          & 13                   & 39                         \\
        \bottomrule
    \end{tabular}
\end{table}

\subsection{Experimental Setup}
\label{experimental-setup}

We implemented the Algorithm \ref{alg:SA} in Python using the Qiskit library
\cite{qiskit}. For each problem size, we ran 5 independent executions of the
VQA and evaluated the mean tour length returned by the best-found circuit,
using 1024 measurement shots to ensure statistical robustness. We also archived
all observed tour lengths, allowing for a direct assessment of the circuit's
ability to produce the exact optimal tour, which was previously determined through
exhaustive enumeration of all permutations.


As a classical baseline, we implemented a SA heuristic capped at $500$ iterations.
The search commenced at a temperature $T$ of $1.0$ and followed a cooling
rate that multiplied the temperature by $0.999$ at each step, terminating when
the temperature dropped below $0.001$ or the iteration limit was reached.

\subsection{Results}

In this section we describe in detail the results with respect to the
experimental method adopted to evaluate the effectiveness of our VQA on the
TSP problem.

The goal is to demonstrate how the proposed method behaves as the problem
size varies (4 to 7 cities) and to quantify the probability of finding the
optimal solution.

\subsubsection{Convergence to the optimal solution}

To evaluate the convergence of our Ansatz optimization method, in this
section we focus on a single training instance. The goal is not to compare all
10 problem settings, but rather to illustrate the behavior of the algorithm
during training and to highlight the effectiveness of our parameter tuning
and $\mathcal{A}$ strategy. This focused analysis demonstrates the efficiency
with which the proposed optimization scheme drives the circuit toward high-quality
solutions.

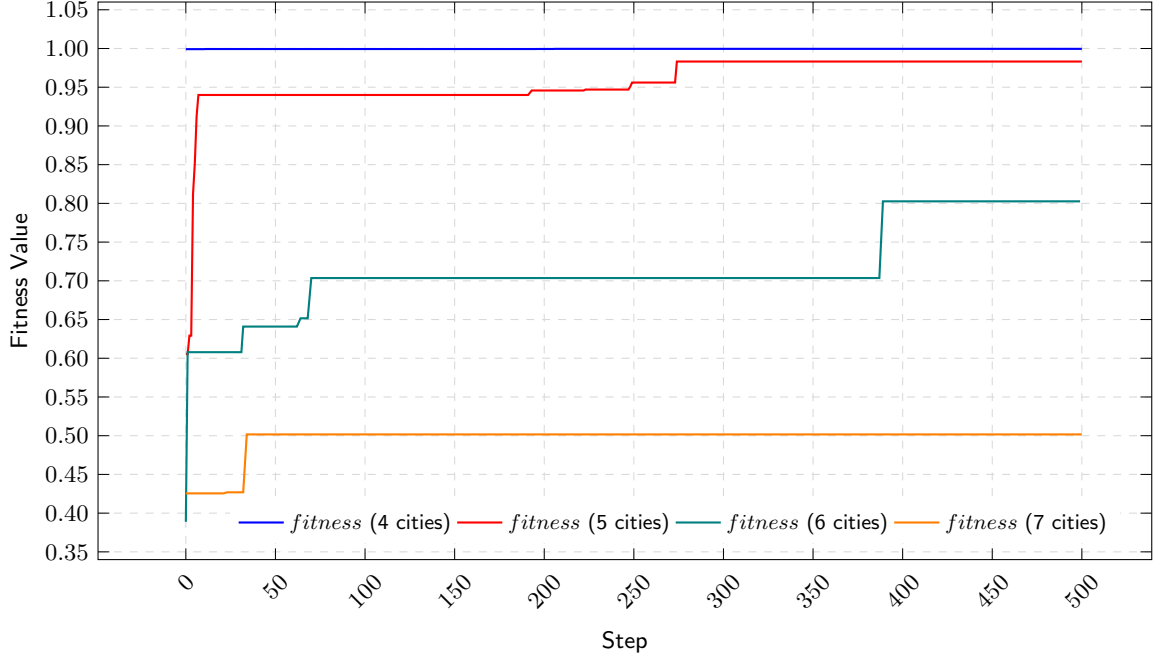
\begin{figure}[ht]
    \centering
    \begin{tikzpicture}
        \begin{axis}[
            matplotlib style,
            xmin=0,
            xmax=490, 
            ymin=0.4,
            ymax=1.0,
            yticklabel style={/pgf/number format/.cd, fixed, fixed zerofill, precision=2},
            xticklabel style={rotate=45},
            ytick distance=0.05,
            enlargelimits=true,
            width=1.0\linewidth,
            height=9cm,
            xlabel=Step,
            ylabel=Fitness Value,
            legend pos=south east,
            legend style={ nodes={scale=0.9, transform shape}, legend columns=-1, },
        ]
            \addplot[ color={blue}, thick, mark=none ] table[x=step, y=value,
            col sep=comma]{best_fitness_4.csv}; \addlegendentry{$fitness$ (4 cities)}

            \addplot[ color={red}, thick, mark=none ] table[x=step, y=value,
            col sep=comma]{best_fitness_5_update.csv}; \addlegendentry{$fitness$ (5 cities)}

            \addplot[ color={teal}, thick, mark=none ] table[x=step, y=value,
            col sep=comma]{best_fitness_6.csv}; \addlegendentry{$fitness$ (6 cities)}

            \addplot[ color={orange}, thick, mark=none ] table[x=step, y=value,
            col sep=comma]{best_fitness_7.csv}; \addlegendentry{$fitness$ (7 cities)}
        \end{axis}
    \end{tikzpicture}
    \caption{Convergence of SA toward the optimal solution probability for
    TSP instances with 4, 5, 6 and 7 cities (index 0, training set). The
    fitness evolves by 500 steps.}
    \label{fig:sa-convergence}
\end{figure}

The trend of $fitness$ that we see in Figure \ref{fig:sa-convergence}
confirms that the algorithm, thanks to the optimized $\mathcal{A}$, quickly
reaches a near-optimal tour and then takes a modest number of steps to fix
its optimality. This result demonstrates the efficiency of the adopted optimization
process.

\paragraph{Best Ansatz found}

\begin{figure*}[ht]
    \centering
    \includegraphics[width=1.0\linewidth]{
        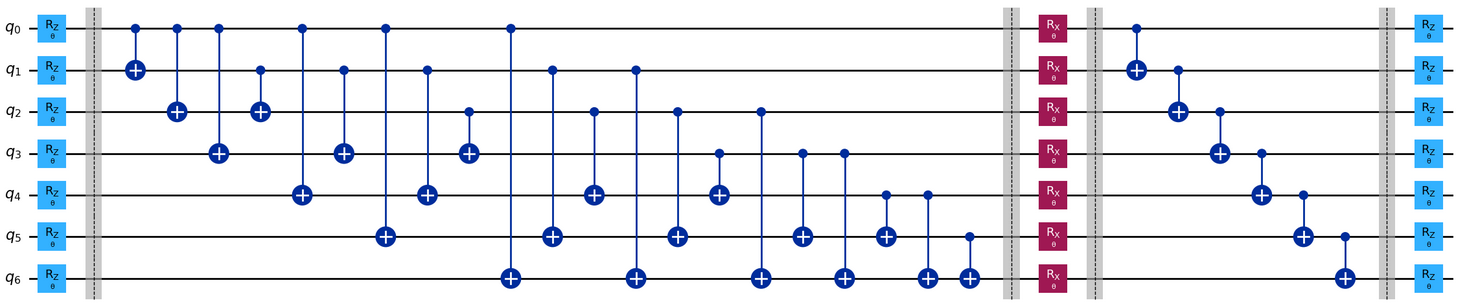
    }
    \caption{Best Ansatz found for training instance with 5 nodes}
    \label{fig:bestAnsatzExample}
\end{figure*}

Figure \ref{fig:bestAnsatzExample} shows the variational circuit that, after
the optimization phase, provided the lowest tour value for the training instance
with 5 cities. The architecture employs all 7 available qubits and is
divided into 3 parametric blocks separated by barriers highlighting the
logical structure.

The optimal $\mathcal{A}$ follows a simple but effective scheme:

\begin{align*}
    \mathrm{R}_{z}\ \text{layer} & \;\longrightarrow\; \text{full entanglement}\;\longrightarrow\; \mathrm{R}_{z}\ \text{layer}    \\
                                 & \;\longrightarrow\; \text{linear entanglement}\;\longrightarrow\; \mathrm{R}_{z}\ \text{layer}.
\end{align*}

Concretely, each qubit is first rotated around the $z$ axis (\textbf{$R_{z}$}).
These single qubit phases are then mixed by a fully connected $CX$ block,
which connects each qubit with all others. A second layer of $R_{x}$ refines
the phases introduced so far, after which a $CX$ scheme provides linear
entanglement. The circuit closes with the $R_{z}$ layer.

\subsubsection{Reusing}

In this section we apply, the optimization structure to unseen problem, and
evaluate the performance of the $\mathcal{A}_{best}$ training. The table \ref{tab:count_multi_threshold}
reports, for each problem dimension, how many of the 10 test TSP instances achieve
an optimal tour sampling probability above a prescribed $\tau$ success threshold.
In the four-city benchmark, the $\mathcal{A}$ generalizes perfectly: all
instances meet even the most stringent requirement, $\tau = 0.90$, leading
to a success rate of $100\%$. When the search space grows to five cities,
the circuit remains highly reliable: 9 of the 10 instances pass each threshold
up to $\tau = 0.80$, while one instance fails to meet the more restrictive
$90\%$ threshold. This shows that even a modest increase in combinatorial complexity
can bring up occasional, but significant, generalization difficulties. The
six-city benchmark highlights the emergence of scalability limitations: with
a relatively permissive threshold ($\tau \le 0.60$), the circuit still
solves six of the ten instances, while imposing more stringent requirements ($\tau
= 0.80$
or $0.90$) reduces the number of successes to four. This result indicates
that, although some six-city instances exceed the circuit's capacity, a
significant portion continue to be successfully solved, signaling good generalization
in many cases.

The deterioration becomes evident as it grows to seven cities. As Tab. \ref{tab:count_multi_threshold}
shows, with a very permissive success threshold, $\tau = 0.20$, the
algorithm manages to sample the optimal tour in only $3$ of the $10$ test
instances. Increasing the threshold to $\tau = 0.40$ reduces the successes to
$2$, and the count remains stuck at $2$ for all subsequent thresholds. In other
words, in $80\%$ times of the seven-city instances the probability of
extracting the optimum remains below $0.60\%$, and in $60\%$ of the cases it
never exceeds $0.20\%$. This sharp decline confirms that the current method
does not efficiently generalize the larger search space created.

\begin{table}[ht]
    \centering
    \setlength{\tabcolsep}{10pt}
    \renewcommand{\arraystretch}{1.0}
    \caption{For each threshold $\tau$, number of test cases ($n_{inst}= 10$)
    in which the optimal tour probability satisfies $P_{opt}\ge \tau$.}
    \label{tab:count_multi_threshold}
    \begin{tabular}{c|cccc}
        \toprule \multirow{2}{*}{\textbf{Threshold $\tau$}} & \multicolumn{4}{c}{\textbf{\#Instances with $P_{opt}\ge\ \tau$}} \\
        \cmidrule(l){2-5}                                   & \textbf{4 Cities}                                               & \textbf{5 Cities} & \textbf{6 Cities} & \textbf{7 Cities} \\
        \midrule 0.20                                       & 10                                                              & 9                 & 6                 & 3                 \\
        0.40                                                & 10                                                              & 9                 & 6                 & 2                 \\
        0.60                                                & 10                                                              & 9                 & 6                 & 2                 \\
        0.80                                                & 10                                                              & 9                 & 4                 & 2                 \\
        0.90                                                & 10                                                              & 8                 & 4                 & 2                 \\
        \bottomrule
    \end{tabular}
    \label{tab:count_multi_threshold}
\end{table}

Therefore, reuse of the $\mathcal{A}$, accompanied by re-optimization of previously
learned parameters, ensures uniform excellence in four-city problems, high but
not perfect robustness in five-city problems, and highly instance-dependent
performance in six-city problems. This behavior provides a clear quantitative
benchmark for identifying instances where better performing circuits or different
optimization strategies will be needed to maintain high performance.

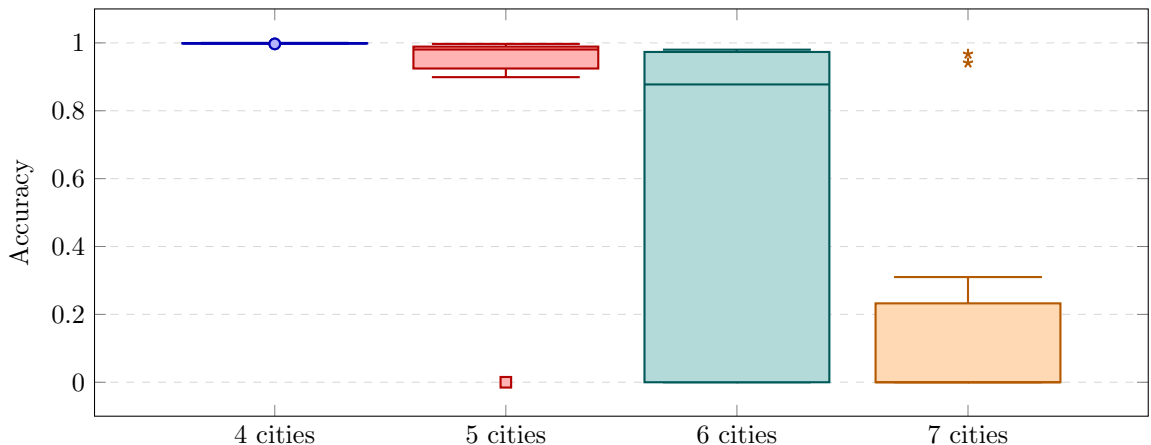
\begin{figure}[htbp]
    \centering
    \begin{tikzpicture}
        \begin{axis}[
            boxplot/draw direction=y,
            ylabel={Accuracy},
            xtick={1,2,3, 4},
            xticklabels={4 cities, 5 cities, 6 cities, 7 cities},
            width=1.0\linewidth,
            height=7cm,
            ymajorgrids,
            grid style={dashed,gray!30},
        ]
            \addplot+[ boxplot, thick, draw=blue!70!black, fill=blue!30, mark
            options={solid},
            ] table[col sep=comma, y index=1]{accuracy_comparison_update.csv};

            \addplot+[ boxplot, thick, draw=red!70!black, fill=red!30, mark
            options={solid},
            ] table[col sep=comma, y index=2] {accuracy_comparison_update.csv};

            \addplot+[ boxplot, thick, draw=teal!70!black, fill=teal!30,
            mark options={solid},
            ] table[col sep=comma, y index=3]{accuracy_comparison_update.csv};

            \addplot+[ boxplot, thick, draw=orange!70!black, fill=orange!30,
            mark options={solid},
            ] table[col sep=comma, y index=4] {accuracy_comparison_update.csv};
        \end{axis}
    \end{tikzpicture}
    \caption{Distribution of solution accuracy across the four problem‑size
    (city‑count) scenarios.}
    \label{fig:accuracy_boxplot}
\end{figure}

In Figure \ref{fig:accuracy_boxplot} we see that, for instances with 4 cities,
the algorithm always returns the optimal tour. All runs coincide with an accuracy
of $1.0$, producing zero variance; the distribution shows that the median,
quartiles and extremes are identical.

For the 5-city instances (red box), the distribution remains narrow and is centered
just below $1.0$. The inter-quartile range (IQR), i.e., the distance between
the third and first quartiles, $Q_{3}- Q_{1}$ \cite{IQR}, extends by a few percentage
points, showing that almost every run reaches the optimal or near-optimal
tour. A single outlier at $0$, however, reveals that the algorithm converges
to a completely wrong minimum.

In contrast, the 6 city instances (green box) show this graph: the lowest $25
\%$ of runs collapse near $0$, while the highest $25\%$ exceeds $0.9$. Although
the median remains high (about $0.9$), the large IQR (about $0.9$) indicates
that performance is highly dependent on instance and initialization, oscillating
between perfect and completely wrong solutions.

For the most challenging instances at $7$ city (orange box in Fig. \ref{fig:accuracy_boxplot})
performance plummets. The median accuracy drops to $\approx 0.20$, with
first quartile $Q_{1}\!\simeq\!0.02$ and third quartile
$Q_{3}\!\simeq\!0.25$, so that the inter-quartile range holds of $0.23\%$.
Both \quotemarks{whiskers} remain anchored at the low end of the axis (whisker
lower than $0$, upper $0.25$) and only two high outliers appear, around $\approx
0.90$. In probabilistic terms, this implies that at least $75\%$ of the runs
sample the optimal tour with probability less than $0.25$, indicating that
the generalization ability of the circuit deteriorates almost completely
when the problem size reaches seven cities.

\section{Conclusions and Future Work}
\label{conclusions_and_future}

This paper introduces an optimization-freezing-reuse study to address TSP with
variational algorithms. By a compact encoding of permutations, the qubit mapping
drops to $\mathcal{O}(n\log n)$, making the representation of TSP instances
much more efficient.

Searching with SA for the optimal $\mathcal{A}$, mutating the circuit topologies
and accepting changes according to the Metropolis criterion, produces after
a hundred iterations a shallow but performant architecture for solving the
problem.

Once the structure is \quotemarks{frozen}, a slight re-optimization of the
parameters by Powell alone sustains impressive performance on unseen
instances, with perfect accuracy for 4 city problems, $90\%$ for 5 cities, and
an average accuracy of $80\%$ for six cities; and, although declining markedly,
an average accuracy of about $20$\% for 7 cities, with rare outliers close
to $0.9\%$, sign that already at this scale the $\mathcal{A}$ shows its limitations.

The big advantage of this method is that, by skipping the costly structural
research in testing, execution times collapse, making the approach
immediately testable on current NISQ devices. However, the very performance
collapse observed at 7 cities indicates that progress is needed on both the
algorithmic and hardware fronts.

The next steps are clear. It will be critical to put it to the test on real quantum
hardware, such as superconducting or ion trap-based hardware. In this context,
it will be imperative to adopt error mitigation techniques, which are essential
to separate the intrinsic performance of the algorithm from the effects of decoherence
and noise to identify the real bottlenecks of the proposed method on real hardware.

From an algorithmic point of view, a natural evolution is to replace the
Powell-based training or testing phase with methods that make use of the
gradient, such as Adam or SPSA. These optimizers could not only speed up
convergence, but also maintain or even improve accuracy as the problem size grows.
In parallel, initializing circuit parameters with warm-start strategies derived
from classical heuristics would allow starting from configurations already close
to good-quality solutions, significantly reducing the training cost as the
number of cities increases.

Finally, it would be extremely interesting to apply the optimize-freeze-reuse
paradigm to even more complex permutation problems, such as the Vehicle
Routing Problem (VRP) and Job-Shop Scheduling (JSS). Both share the burden of
exploring an endless combinatorial space, but introduce additional
constraints that can be handled with penalty terms or, integrated directly into
the $\mathcal{A}$. Demonstrating the robustness of the method on VRP and JSS
would not only prove the usefulness and operation of the method, but also pave
the way for the application of the workflow to a whole class of NP-hard problems
of great industrial relevance.

\bibliographystyle{ieeetr} 
\bibliography{paper}

\end{document}